\documentclass[a4paper,twoside]{article}

\usepackage{epsfig}
\usepackage{subcaption}
\usepackage{calc}
\usepackage{amssymb}
\usepackage{amstext}
\usepackage{amsmath}
\usepackage{amsthm}
\usepackage{multicol}
\usepackage{pslatex}
\usepackage{apalike}
\usepackage{algorithm2e}
\usepackage[bottom]{footmisc}
\usepackage{SCITEPRESS}
\usepackage{makecell}
\usepackage{url}
\usepackage{xcolor}
\usepackage{hyphenat}

\begin{document}

\title{Identifying and Understanding Human Values in Text: \\A Tailorable \textit{LLM}-based Architecture}

\author{\authorname{Eduardo de la Cruz\sup{1,2,}\thanks{Corresponding author: eduardo.cruz@urjc.es}\orcidAuthor{0009-0009-2691-4330}, Marcelo Karanik\sup{2}\orcidAuthor{0000-0001-8848-3681} and Sascha Ossowski\sup{2}\orcidAuthor{0000-0003-2483-9508}}
\affiliation{\sup{1}Universidad Politécnica de Madrid, Madrid, Spain}
\affiliation{\sup{2}CETINIA, Universidad Rey Juan Carlos, Madrid, Spain}
\email{\{eduardo.cruz, marcelo.karanik, sascha.ossowski\}@urjc.es}}


\abstract{
As intelligent systems become more autonomous, the scientific community focuses on creating decision-making mechanisms that include ethical and moral considerations, unlike traditional utility-maximisation models. To achieve this, a key aspect is assessing how well these decisions align with human values. To this end, a promising line of research is centred on developing approaches based on Large Language Models (\textit{LLM}s) to identify human values from text, whether explicit or implicit, enabling their recognition throughout. This paper introduces a \textit{LLM}-based architecture to detect and quantify the intensity of human values in text, avoiding the limitations of previous approaches tied to specific value theory or complex prompt engineering. The architecture comprises three coordinated modules: one that generates structured value specifications from the foundational texts of any theoretical framework; one that labels texts using these specifications; and one that assigns graded support or resistance based on rhetorical and semantic evidence. This modular approach separates the tasks of conceptualising from detecting human values, creating a scalable and reproducible process driven by value specifications adaptable to various theories. The architecture was instantiated with multiple \textit{LLM}s and evaluated using the ValueEval dataset. The experiments demonstrate good detection performance, confirming the generality of the pipeline.}

\onecolumn \maketitle \normalsize \setcounter{footnote}{0} \vfill

%%%%%%%%%%%%%%%%%%%%%%%%%%%%%%%%%%%%%%%%%%%%%%%%%%%%%%%%%%%%%%%%%%%%%%%%

\section{\uppercase{Introduction}} \label{intro}
Decision-making, both human and algorithmic, is directly influenced by the expected benefit of each available option, but other, often subjective, factors also play a crucial role. These include preferences for particular tangible or intangible characteristics of each option, as well as the ethical and moral considerations of the decision-maker. Under specific circumstances, an agent may therefore choose an option that does not maximise utility but better aligns with their values. For instance, some people might decide not to buy a cheap and polluting car and instead opt for a more expensive, environmentally friendly vehicle, reflecting a commitment to environmental sustainability.

This topic has been extensively studied in philosophy and psychology \cite{Maio2010,maslow:motivation,Schwartz2012Refining}, giving rise to various theories that define human values, their traits, and their interactions. These theories aim to understand how human values influence decisions across different domains \cite{SAATY2010963}.

Building on these theories, recent advances in intelligent systems capable of autonomous decision-making have prompted researchers to reassess the decision models employed in such systems. With the significant progress of artificial intelligence, designing autonomous systems that are aware of human values has become a priority for the scientific community \cite{osman2023computational}.

For an autonomous system to be value-aware, its decisions must align with a set of values \cite{Montes2021,serramia2020} that, in addition to maximising utility, incorporate ethical and moral considerations. A central task in the value alignment process is to determine which values each available action promotes or demotes. While some works base this alignment on static considerations defined during system design, the current challenge is to detect these values automatically in a way that is systematic, replicable, and flexible.

A promising approach is to generate textual descriptions of the objectives and implications of each action. Natural language processing (NLP) techniques can then be applied to these descriptions to extract the embedded values and analyse how each action affects them.

This article presents a three-module architecture that formalises the process of value analysis in text. Its key innovation is a theory-agnostic capability, enabled by the Value Conceptualisation Module (VCM), which uses large language models (\textit{LLM}s) to analyse a given value theory from academic documents. From this conceptualisation, structured specifications are generated that guide the detection module in identifying the presence of values in a given text, while an intensity scale models the degree to which each value is promoted or demoted. The architecture orchestrates a complete workflow from theoretical conceptualisation to intensity rating, moving beyond approaches based solely on prompt engineering. The model has been implemented using \textit{Llama-4-scout}, \textit{Gemma3}, \textit{DeepSeek-R1}, and \textit{Qwen3}, which have shown closely similar performance in the experiments. A reference implementation is publicly available\footnote{\url{https://huggingface.co/spaces/segoedu/valuelens}}.

The rest of this article is organised as follows. Section~\ref{RWork} reviews related work, providing context for the model described in Section~\ref{architecture}. Section~\ref{sim&res} details the simulations performed and the results obtained. Section~\ref{conclusions} discusses the main considerations related to the proposed model and outlines potential future work.

%%%%%%%%%%%%%%%%%%%%%%%%%%%%%%%%%%%%%%%%%%%%%%%%%%%%%%%%%%%%%%%%%%%%%%%%

\section{\uppercase{Related work}}\label{RWork}
Human values are intrinsic to human beings: they guide behaviour and play a crucial role in decision-making. There is broad agreement in the social sciences that values represent beliefs about desirable states or outcomes and shape both personal and societal behaviour patterns \cite{guth1965,rokeach1967survey,SCHWARTZ19921,maslow:motivation,Maio2010}.

On this basis, several theories define the structure of human values and their interactions. Among the most influential are Moral Foundations Theory, which proposes six universal moral values \cite{Haidt13}, and the Basic Human Values framework (Schwartz's Value Theory), which evolves from ten to nineteen values and analyses their relationships in terms of motivational compatibility and incompatibility \cite{SCHWARTZ19921,Schwartz2012Refining}. These theories provide the conceptual foundations for computational models of decision-making.

Such value theories have inspired numerous studies that incorporate values into the decision-making processes of autonomous systems, including value-based argumentation frameworks \cite{vanderWeide10}, formal reasoning frameworks \cite{Steels24}, and, more recently, the use of \textit{LLM}s to generate value-aligned responses \cite{Giulio24}. In AI engineering, this has led to a focus on value-aware autonomous agents and two main challenges: how to represent values and their interactions, and how to ensure that decisions align with them. Approaches include taxonomies linked to computational concepts \cite{osman2023computational,kiesel-etal-2022-identifying}, explicit preference relations over values \cite{Pommeranz2012,Liao2019,Siebert2022}, and fuzzy measures for individual and combined values \cite{karanik-ACM}, as well as participatory evaluation, value-aligned states and norms, and aggregation methods for alignment \cite{Liscio2022,Siebert2022,serramia2020,Montes2021,Lera-leri,karanik-ACM}.

A key open issue is to determine which values are promoted or demoted by a given action before it is executed. Existing models often rely on static promotion schemes derived from value definitions and designer judgement, which become inadequate when contextual changes alter which values are actually promoted \cite{karanik-ACM,karanik-sinergy}. This limitation motivates methods that infer value promotion and demotion dynamically from textual descriptions of actions. Classical NLP techniques can be applied to such descriptions, ranging from word-counting approaches \cite{Fulgoni16} to embedding-based methods \cite{Kennedy21} and lexicons focused on moral vocabulary \cite{Mokhberian_2020,Hopp}.

With the progress of generative artificial intelligence, detecting human values in text has emerged as a central research topic. Early work relied on transformer-based models and ensembles (e.g., BERT, RoBERTa, DeBERTa) to classify values in argumentative texts, often in shared tasks such as SemEval, and explored specialised attention mechanisms with contrastive learning \cite{schroter2023adamsmithsemeval2023task4,saha2023rudolfchristopheuckensemeval2023,fang2023epicurussemeval2023task4,zhang2023maozedongsemeval2023task4}. More recent studies leverage \textit{LLM}s through prompt engineering and supervised fine-tuning (SFT), investigating prompting schemes (zero-shot, few-shot, chain-of-thought), the moral reasoning capabilities of \textit{LLM}s, and the values they express in real-world interactions \cite{SENTHILKUMAR2025,khamassi2024strong,BULLA2025100609,huang2025}. Comparisons between prompting and SFT across models such as GPT-4, Llama, and Gemini highlight trade-offs in performance, data requirements, and flexibility \cite{MISHRA2024}.

In parallel, several works explore modular architectures for value detection. EAVIT combines a local, fine-tunable \textit{LLM} for initial detection with a larger online \textit{LLM} for final identification \cite{zhu2025}, while Value Lens proposes a two-stage \textit{LLM}-based pipeline that first formalises a value theory and then detects values in text \cite{delacruz2025valuelens}, within a broader trend towards modular AI systems \cite{delacruz2024ia}. Unlike these approaches, the framework proposed in this article derives structured value specifications directly from foundational texts, reducing reliance on ad-hoc prompt engineering and theory-specific implementations, and uses them as the basis for value detection in text.

%%%%%%%%%%%%%%%%%%%%%%%%%%%%%%%%%%%%%%%%%%%%%%%%%%%%%%%%%%%%%%%%%%%%%%%%

\section{\uppercase{Proposed Architecture}} \label{architecture}
\begin{figure*}
    \centering
    \includegraphics[width=\textwidth]{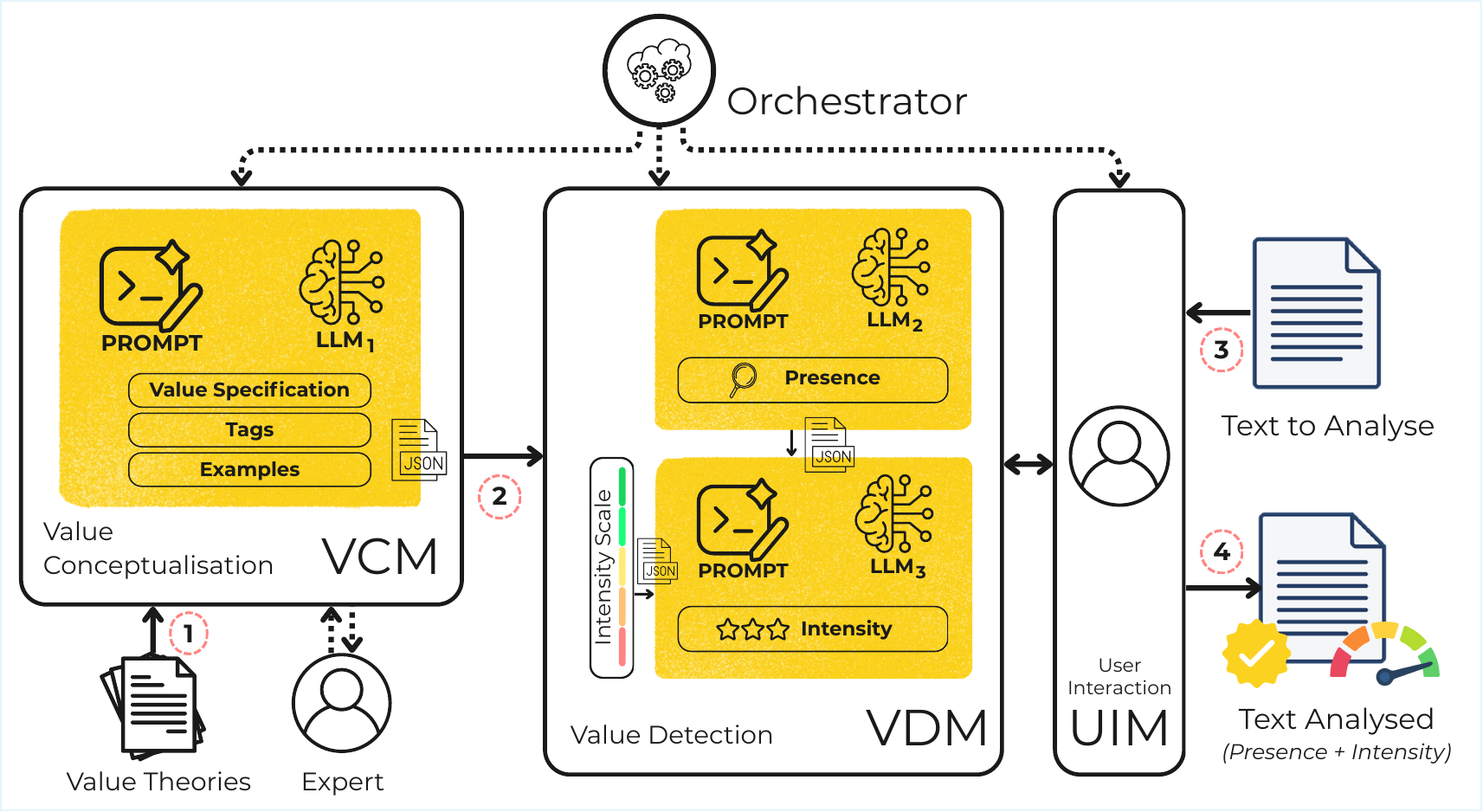}
    \caption{Value detection architecture.}
    \label{fig:arch}
\end{figure*}

The proposed architecture for detecting the presence and intensity of human values in text comprises three main modules, designed to ensure precision, modularity, and robustness. As shown in Figure~\ref{fig:arch}, the first module is the \textit{Value Conceptualisation Module} (VCM), the second is the \textit{Value Detection Module} (VDM), and the third is the \textit{User Interaction Module} (UIM). Extending recent multi-stage pipelines like \textit{Value Lens} \cite{delacruz2025valuelens}, a central orchestrator manages the interactions between these modules, elevating the solution from a collection of prompts to a coherent and automated system that handles the complete information flow. In the following subsections, the model's functionality is explained in more detail, describing each module and its orchestration.

To illustrate the architecture's functionality, each module description includes a case where the user aims to identify which values from Schwartz's value theory \cite{SCHWARTZ19921,Schwartz2012Refining} and their corresponding intensities are present in the following text: \textit{Climbing the corporate ladder used to be my goal, but I've realised that personal fulfilment matters more than titles or paychecks. Success is now about balance and happiness.}

\subsection{Value Conceptualisation Module} \label{VCM}
Value conceptualisation involves understanding the key elements of a reference value theory to serve as a basis for identifying values in text. This module ensures the architecture's flexibility by processing foundational texts (e.g., academic papers, official descriptions) to automatically generate machine-interpretable specifications, allowing adaptation to new frameworks without re-engineering.

The VCM (Figure~\ref{fig:arch}) employs a large language model, $\text{\textit{LLM}}_1$, utilizing a knowledge transfer prompt to extract and structure this information. The names, descriptions, groupings, tags, and examples of each value are stored in a JSON file. Optionally, experts may inspect and modify these specifications to enhance their quality and consistency, while the system remains fully autonomous. Finally, this enriched JSON description is forwarded to the detection module.

This conceptualisation process supports both one-time setups and continuous improvement. The input from new documents or expert feedback enables ongoing enhancements that refine the value theory definitions, serving as ground truth for the detection module.

Regarding the running example, the documents explaining Schwartz's theory are processed by $\text{\textit{LLM}}_1$ to extract value specifications, tags and examples. This results in a list of nineteen values; for brevity, Table~\ref{tab:valconcep} displays the conceptualisation for the two values most relevant to the subsequent detection analysis.

\begin{table*}[t]
    \fontsize{9pt}{10pt}\selectfont
    \caption{Schwartz value theory conceptualisation.}
    \label{tab:valconcep} 
    \centering
    % La segunda columna se calcula: Ancho total - 2cm - espacios de padding y líneas
    \begin{tabular}{|p{1.9cm}|p{\dimexpr\textwidth-2cm-4\tabcolsep-3\arrayrulewidth\relax}|}
        \hline
        \textbf{Value ID} & \textbf{Conceptualisation Tags \& Examples} \\
        \hline
        : & : \\
        \hline
        Power (P) & \textbf{Tags:} Power tags \textbf{Examples:} Power examples\\
        \hline
        Achieve-ment (ACH) & 
        \textbf{Tags:} success, competence, ambition, goal-oriented, excellence, recognition, winning, accomplishment, status, performance, determination, productivity, prestige, leadership, competitiveness \newline
        \textbf{Examples:} Winning awards; Getting a promotion; Achieving high grades; Being recognised for accomplishments; Meeting challenging goals; Completing a marathon; Receiving public praise; Graduating with honours; Finishing a big project; Being top of your class; Earning a certification; Setting and achieving personal records; Being elected as a leader; Launching a business; Overcoming challenges \\
        \hline
        Self-Direction (SDI) & 
        \textbf{Tags:} independence, creativity, freedom, curiosity, autonomy, exploration, innovation, originality, open-mindedness, initiative, critical thinking, self-expression, problem-solving \newline
        \textbf{Examples:} Thinking independently; Pursuing creative hobbies; Making your own decisions; Exploring new ideas; Learning new skills; Expressing your opinions; Starting your business; Travelling alone; Choosing your own career path; Inventing something new; Reading to expand knowledge; Questioning conventional wisdom; Writing stories; Developing unique solutions; Trying new technologies \\
        \hline
        Univer.(U) & \textbf{Tags:} Universalism tags \textbf{Examples:} Univeralism examples\\
        \hline
        : & : \\
        \hline
    \end{tabular}
\end{table*}

\subsection{Value Detection Module - VDM} \label{sec:vdm}
The Value Detection Module (VDM) (Figure~\ref{fig:arch}) performs two main tasks: first, it identifies which human values are expressed in a text; second, estimates the intensity with which each value is promoted or demoted. This separation between detection and rating improves analytical precision and facilitates the interpretation of how values are prioritised in the text.

The analysis carried out by the VDM is grounded in the value theory conceptualised by the VCM (Figure~\ref{fig:arch}) and relies on two large language models, $\text{\textit{LLM}}_2$ and $\text{\textit{LLM}}_3$. $\text{\textit{LLM}}_2$ is responsible for detecting values, while $\text{\textit{LLM}}_3$ assigns intensity scores and generates supporting justifications.

For value detection, $\text{\textit{LLM}}_2$ receives the JSON-encoded enriched value specifications produced by the VCM, together with the input text. A second knowledge transfer prompt instructs the model to identify both explicit and implicit references to the values defined in the theory and to assess their relevance for the overall meaning and motivation of the text. As a result, the text is labelled with a list of values that are considered present. 

For the running example, the list of detected values is shown in Table~\ref{tab:valdetected}, obtained with $\text{\textit{LLM}}_2$ set as \textit{Llama4} (temperature 0.0).

\begin{table}[ht]
    \fontsize{9pt}{10pt}\selectfont
    \caption{Value detection list for the sample text.}
    \centering
    \begin{tabular}{|p{1.85cm}|p{4.65cm}|}
        \hline
        Value ID & Key Evidence in Text \\
        \hline
        Achievement (ACH) &``\textit{climbing the corporate ladder used to be my goal''} \\
        \hline
        Self-Direction (SDI) &``\textit{personal fulfillment matters more''}, ``\textit{balance and happiness''}\\
        \hline
    \end{tabular}
    \label{tab:valdetected}
\end{table}

While recognising which values are present is informative, it does not provide sufficient detail about the strength of each identified value's relationship with the text, which is crucial for value alignment and subsequent decision-making. To address this, a third knowledge transfer prompt is used with $\text{\textit{LLM}}_3$ to assess the intensities of the detected values, using the intensity scale described in Table \ref{tab:valuestances}. 

Once the detection process is complete, the VDM produces a structured JSON output that links each detected value to an intensity label and a concise justification. 

\begin{table}[ht]
    \centering
    \caption{Value Intensity Scale}
    \label{tab:valuestances}
    % Restamos el padding y los bordes del ancho de línea
    \begin{tabular}{|p{\dimexpr\linewidth-2\tabcolsep-2\arrayrulewidth}|}
        \hline
        \textbf{(+ + +) Strong support}: The text fervently promotes and defends the value, emphasising its importance. This value is central to the message, backed by emotional, moral, and logical urgency. \\
        \hline
        \textbf{(+) Mild support}: The text aligns with the value through positive mention or subtle endorsement, without significant detail, insistence, or emphasis. \\
        \hline
        \textbf{(o) Neutral}: The text presents the value neutrally without showing clear support or opposition. The tone is factual, balanced, and incidental. \\
        \hline
        \textbf{(--) Mild resistance}: The text subtly questions, downplays, or presents alternative perspectives on its value. This opposition is indirect, cautious, or expressed through mild scepticism. \\
        \hline
        \textbf{(-- -- --) Strong resistance}: The text challenges, criticises, or undermines its value directly and forcefully. This includes explicit arguments, a negative emotional tone, or repeated rejections. \\
        \hline
        \textbf{($\pm$) Reframing}: The text acknowledges its value but shifts its meaning and context, introducing a new perspective that changes the emphasis without openly expressing support or opposition. \\
        \hline
        \textbf{($\varnothing$) No values}: The text is factual or technical in nature and does not contain evaluative statements. \\
        \hline
    \end{tabular}
\end{table}

Returning to the example, $\text{\textit{LLM}}_3$ receives the enriched descriptions of the values together with the intensity scale and outputs the list of values with their corresponding intensities and explanations, as shown in Table~\ref{tab:output}. These components allow the VDM to present a comprehensive perspective on the insights related to the values in the text.

\begin{table}[ht]
    \fontsize{9pt}{10pt}\selectfont
    \caption{Model output for the sample text.}
    \centering
    \begin{tabular}{|p{1.2 cm}|p{1.0cm}|p{4.1 cm}|}
        \hline
        Value ID & Intensity & Justification \\
        \hline
        Achieve-ment (ACH) & Mild resistance (--)&  While the text mentions a desire to ``climb the corporate ladder'' it frames this as a former goal that has been superseded by a focus on personal fulfillment. This suggests a shift away from achievement-oriented values. \\
        \hline
        Self-Direction (SDI) & Strong support (+ + +)& The text explicitly states that ``personal fulfilment matters more than titles or paychecks'' and defines ``success'' as ``balance and happiness'' prioritising personal growth and autonomy over external achievements. \\
        \hline
    \end{tabular}
    \label{tab:output}
\end{table}

\subsection{User Interaction Module - UIM} \label{UIM}
The UIM acts as the central hub for the system, facilitating input management and result interpretation. Through this interface, users upload the text to analyse and select the target value theory (Figure~\ref{fig:arch}). Crucially, the module enables expert inspection and modification of the VCM-generated specifications prior to analysis, ensuring the theoretical definitions are accurate before the detection phase begins.

Once processed by the VDM, the interface goes beyond merely listing labels. It visualizes the consolidated results generated by $\text{\textit{LLM}}_2$ and $\text{\textit{LLM}}_3$, displaying the textual evidence, specific justifications, and the intensity indicators for each value. These visual cues allow users to quickly evaluate the text's alignment---identifying the degree of value promotion or demotion---without parsing raw data. It should be clarified that the graphical interface includes additional visual features to assist in understanding the detection for each identified value. Regarding the running example, the UIM displays the final output shown in Table~\ref{tab:output}, augmented with these graphical aids to facilitate interpretation.

\subsection{Orchestration} \label{Orch}
The three modules described above have distinct objectives, but their integration requires a coordination mechanism. The orchestrator fulfils this role by managing the interactions between modules and ensuring that the value detection pipeline operates as a coherent, automated workflow.

To clarify the orchestration process (Figure~\ref{fig:arch}), note that the conceptualisation of the value theory (VCM activity) operates independently of value detection (VDM activity) and user interaction (UIM activity). The orchestrator monitors the repository of foundational documents and triggers the VCM to update value specifications via flow control~1 whenever changes are detected, ensuring background updates do not interfere with ongoing analysis requests.

When a new analysis request is received, the orchestrator initiates the second main activity: value detection. It retrieves the enriched value specifications from the VCM and forwards them to the VDM (flow control~2), together with the text to be analysed (flow control~3). The VDM then returns the labelled text, including value intensities and explanations, and the orchestrator delivers these results to the user through the UIM (flow control~4), enabling users to assess how the text aligns with each detected value.

%%%%%%%%%%%%%%%%%%%%%%%%%%%%%%%%%%%%%%%%%%%%%%%%%%%%%%%%%%%%%%%%%%%%%%%%

\section{\uppercase{Simulations and results}} \label{sim&res}
This section describes the experimental design for evaluating the performance of the proposed architecture in detecting and analysing human values in text. It includes details about the hardware configuration, the selection of \textit{LLM}s, and presents the results obtained. The findings demonstrate the robustness of the architecture, showing that the methodological framework is more decisive for performance than the specific \textit{LLM} choice.

\subsection{Experiment Setup}
The experiment aims to evaluate the effectiveness of various \textit{LLM}s in recognising human values as outlined by Schwartz’s value theory \cite{Schwartz2012Refining}. This theoretical framework was selected due to its widespread acceptance in psychology and the availability of a relevant text dataset, which enables accurate performance measurement and comparison.

The simulation process is based on the value detection module described in Section~\ref{sec:vdm}. The Touché24 - ValueEval dataset \cite{touche24valueval}, containing 59,662 short texts each labelled with one or more values, was employed. For the simulations, a subset of 7,600 texts was selected, balancing statistical reliability with computational feasibility and enabling systematic comparison across multiple model configurations. 

To ensure an unbiased evaluation, the labels from these texts were removed before being passed as input to the $\text{\textit{LLM}}_2$. The model's task was to generate a labelled dataset by linking the appropriate values from Schwartz's theory to each text sample.

The performance of the models was subsequently assessed using micro \textit{F1-score}, \textit{precision}, and recall metrics, which were determined by comparing the $\text{\textit{LLM}}_2$-generated labels to the original annotations in the Touché24-ValueEval dataset.

\subsection{Hardware and \textit{LLM} Selection}
Simulations were conducted on a local machine equipped with an NVIDIA H100 GPU containing 96 GB of VRAM. The software environment consisted of a local Python 3 installation, with an Ollama server for model deployment and inference. To ensure reproducibility, all models were run with a temperature set to 0.0 and a seed fixed at 42. For comparative purposes, the \textit{Gemma3} model was also evaluated at a temperature of 1.0 to observe its effect on performance. To analyse how performance on the value detection task is influenced by different model architectures and parameter sizes, a diverse set of open-weight \textit{LLM}s was selected, all employing Q4\_K\_M quantisation. The chosen models were \textit{Gpt-oss} (120B), \textit{Llama-4-scout} (109B), \textit{Qwen3} (32.8B), \textit{DeepSeek-R1-Distill-Qwen-32B} (32.8B), and \textit{Gemma3} (27.4B).

\subsection{Performance Analysis}\label{res}
The performance of each model was evaluated using the micro \textit{F1-score}, \textit{recall}, and \textit{precision}. The results are displayed in the Table~\ref{tab:performance}.

\begin{table}[ht]
    \fontsize{9pt}{10pt}\selectfont
    \caption{Model Performance Comparison.}
    \centering
    \begin{tabular}{|l|c|c|c|}
        \hline
        Model & Micro \textit{F1-score} & \textit{Recall} & \textit{Precision} \\
        \hline
        \textit{Gemma3}           & 0.3406 & 44.8\% & 27.5\% \\
        \hline
        \textit{Gpt-oss}          & 0.3359 & 33.2\% & 34.0\% \\
        \hline
        \textit{Llama4-scout}     & 0.3275 & 48.1\% & 24.8\% \\
        \hline
        \textit{DeepSeek--R1}     & 0.3227 & 30.2\% & 34.7\% \\
        \hline
        \textit{Qwen3}            & 0.3216 & 27.5\% & 39.1\% \\
        \hline
    \end{tabular}
    \label{tab:performance}
\end{table}

\textbf{Micro \textit{F1-score}}. The micro \textit{F1-score} is the harmonic mean of \textit{precision} and \textit{recall}, computed globally by counting the total true positives, false negatives, and false positives, which offers a balanced assessment of model performance. 

The results indicate that the \textit{Gemma3} model achieved the highest score (0.3406). However, the key finding is the minimal performance difference among the high-end models. The \textit{Gpt-oss} (0.3359), \textit{Llama4-scout} (0.3275), \textit{DeepSeek-R1} (0.3227), and \textit{Qwen-3} (0.3216) models all scored within a very narrow range.
This tight clustering of scores among diverse \textit{LLM}s demonstrates the effectiveness of the proposed process: it can be effectively tailored to the use of different models without significant loss of accuracy, and its best micro F1-score is comparable to classical multi-label value detection baselines reported on the ValueEval contest \cite{touche24valueval}. 

\textbf{\textit{Precision} and \textit{Recall}}. \textit{Precision} measures the proportion of true positive predictions among all positive predictions, indicating the accuracy of the model's detections. \textit{Recall} quantifies the proportion of actual positives that were correctly identified, reflecting the model's completeness. An analysis of these metrics reveals a clear trade-off among the models, which directly impacts their \textit{F1-scores}.

\textit{Qwen3} achieved the highest \textit{precision} at 39.1\%, indicating its value detections were the most likely to be correct, but this came with a low \textit{recall} of 27.5\%. Conversely, \textit{Llama4-scout} registered the highest \textit{recall} at 48.1\%, showing it was the most effective at identifying all relevant values, though at the cost of lower \textit{precision} (24.8\%). \textit{DeepSeek‑R1} reports 34.7\% \textit{precision} and 30.2\% \textit{recall}, reflecting a modest emphasis on correctness over coverage. In contrast, \textit{Gpt‑oss} is more even across both dimensions, with 34.0\% \textit{precision} and 33.2\% \textit{recall}, positioning it as one of the most stable performers across both metrics.

This inherent tension is critical because the \textit{F1-score}, as the harmonic mean, is penalised by a significant imbalance between \textit{precision} and \textit{recall}. Consequently, a model with more harmonised metrics, even if neither is the absolute highest, can achieve a superior overall \textit{F1-score}.

\textbf{Temperature}. In \textit{LLM}s, temperature is a hyperparameter that controls the level of randomness in the generated output, striking a balance between predictable and creative responses. It functions by scaling the probability distribution of potential next tokens, where a lower temperature (e.g., 0.0) makes the model more deterministic and a higher temperature (e.g., 1.0) increases variability. The seed is used to ensure that this randomness is reproducible across runs (e.g., T=1.0, S=42 will always produce the same output), whereas a non-fixed seed would produce different results each.

The evaluation of the \textit{Gemma3} model at different temperatures (i.e., 1.0 and 0.0) showed a negligible impact on performance (see Table~\ref{tab:gemma3_performance}). 

\begin{table}[ht]
    \fontsize{9pt}{10pt}\selectfont
    \caption{\textit{Gemma3} Performance at Different Temperatures}
    \centering
    \begin{tabular}{|l|c|c|c|}
        \hline
        Temperature & Mic. \textit{F1-score} & \textit{Recall} & \textit{Precision} \\
        \hline
        (T=0.0) & 0.3406 & 44.8\% & 27.5\% \\
        \hline
        (T=1.0) & 0.3414 & 44.9\% & 27.5\% \\
        \hline
        (T=1.0, S=42) & 0.3407 & 44.7\% & 27.5\% \\
        \hline
        (T=1.0, S=123) & 0.3391 & 44.6\% & 27.4\% \\
        \hline
    \end{tabular}
    \label{tab:gemma3_performance}
\end{table}

The Micro \textit{F1-score}, \textit{Recall}, and \textit{Precision} metrics showed a striking similarity across both configurations, demonstrating a high degree of consistency in performance. This stability strongly suggests that the carefully crafted and restrictive prompts in the proposed architecture effectively constrained the models' output, minimising the creative variability that temperature changes typically introduce.

This indicates that the \textit{LLM}'s responses were heavily guided by the structure and specificity of the prompts, rather than influenced by randomness or variance in output generation. By clearly specifying the structure and format of the expected response, the prompts effectively limited the potential range of valid next tokens to a very narrow set of options. This specificity directed the model's focus towards producing responses that adhered closely to a predetermined template. Consequently, the nature of the task shifted from one of creative generation, where diverse and imaginative outputs could emerge, to a more deterministic approach, where the emphasis was on strict compliance with the outlined criteria. This shift not only influenced the creativity of the generated responses but also ensured a more uniform outcome, as the model concentrated on fulfilling the outlined requirements rather than exploring a broader range of possibilities. In such a constrained environment, the probability of the required tokens is so dominant that altering the temperature, which modulates the likelihood of less probable tokens, has little to no effect on the final output.

%%%%%%%%%%%%%%%%%%%%%%%%%%%%%%%%%%%%%%%%%%%%%%%%%%%%%%%%%%%%%%%%%%%%%%%%

\section{\uppercase{Conclusions}}\label{conclusions}
This article presents a comprehensive and agnostic architecture for detecting and identifying human values in textual descriptions, designed to overcome the limitations of ad-hoc prompt-based approaches. The model is structured into three interconnected modules: the Value Conceptualisation Module (VCM), which defines values based on established value theories and can flexibly adapt to any framework by processing foundational texts; the Value Detection Module (VDM), which leverages \textit{LLM} techniques to identify values and their nuances within text; and the User Interaction Module (UIM), which enables exploration of the detected values and provides insights. While fully automated, the architecture supports human-in-the-loop intervention to refine value specifications.

The contribution of this architecture lies in its modularity, which ensures a clear separation of concerns. This enhances tailorability to diverse value theories and overcomes the limitations of ad-hoc prompt-based approaches for reliably analysing human values in texts. As lessons learnt from this work, it is highlighted that, rather than searching for a ``perfect prompt'', models for value-aligned autonomous decision-making should rely on principled processes and structured architectures. The model presented in this paper is one step towards this goal. Furthermore, the study concludes that effective architectural design is a more crucial factor than the subtle differences between leading models.

The present results are promising, yet future work could profitably include ablation studies on the architecture, experiments with additional value theories, human validation of the proposed intensity scale, and evaluations on a broader range of texts and use cases. Complementing these directions, ongoing work explores both the integration of this architecture into a broader value-based decision model and the estimation of value promotion and demotion from partial information, following Schwartz’s theory of values.

%%%%%%%%%%%%%%%%%%%%%%%%%%%%%%%%%%%%%%%%%%%%%%%%%%%%%%%%%%%%%%%%%%%

\section*{\uppercase{Acknowledgements}}
Supported by grants VAE: TED2021-131295B-C33 funded by MCIN/AEI/10.13039/501100011033 and EU NextGenerationEU/PRTR, and EVASAI: PID2024-158227NB-C32 funded by MICIU/AEI/10.13039/501100011033/FEDER, UE.

%%%%%%%%%%%%%%%%%%%%%%%%%%%%%%%%%%%%%%%%%%%%%%%%%%%%%%%%%%%%%%%%%%%

\bibliographystyle{apalike}
{\small
\bibliography{bibliography.bib}}

\end{document}